%% file: main.tex
\documentclass[11pt]{article}

\usepackage[final]{acl}

\usepackage{times}
\usepackage{latexsym}

\usepackage{xspace}
\usepackage{subcaption}
\usepackage[T1]{fontenc}

\usepackage[utf8]{inputenc}

\usepackage{microtype}

\usepackage{inconsolata}

\usepackage{graphicx}
\usepackage{amsmath}
\usepackage{booktabs}
\usepackage{multirow}

\usepackage[dvipsnames]{xcolor}

\usepackage{mystyle}
\usepackage{mlsymbols}

\input{latex/common}
\title{Beyond Supervised Clarification: Input Rewriting with LLMs \\ for Dialogue Discourse Parsing}

\author{
 \textbf{Yiming Liu\textsuperscript{1}},
 \textbf{Ziyue Zhang\textsuperscript{1}},
 \textbf{Zhichao Xu\textsuperscript{2}},
 \textbf{Xin Yu\textsuperscript{2}},
\\
 \textbf{Yingheng Tang\textsuperscript{3}},
 \textbf{Tianyu Jiang\textsuperscript{4}},
 \textbf{Jie Cao\textsuperscript{1}}
\\
 \textsuperscript{1}University of Oklahoma,
 \textsuperscript{2}University of Utah \\
 \textsuperscript{3}Lawrence Berkeley National Laboratory,
 \textsuperscript{4}University of Cincinnati
\\
\small{
  \texttt{\{ymliu,ziyue.zhang-1,jie.cao\}@ou.edu,}
  \texttt{\{zhichao.xu,xin.yu\}@utah.edu}
  } \\
\small{
  \texttt{ytang4@lbl.gov,}
  \texttt{tianyu.jiang@uc.edu}
  }
}

\begin{document}
\maketitle

\input{latex/0_abstract}
\input{latex/1_intro}

\input{latex/2_related}
\input{latex/4_methods}

\input{latex/5_analysis}
\input{latex/6_discussion}
\input{latex/7_conclusion}
\input{latex/8_limitations}
\input{latex/9_ack}
\bibliography{custom}

\appendix
\input{latex/app-dataset}
\input{latex/app-prompts}

\input{latex/app-rl}
\input{latex/app-parsers}
\input{latex/app-analysis}

\input{latex/app-qual}
\end{document}

%% file: latex/common.tex
\definecolor{mdgreen}{rgb}{0.05,0.6,0.05}
\definecolor{darkcyan}{rgb}{0.0, 0.55, 0.55}
\newcommand{\lbl}[1]{\textsc{#1}}

\newcommand{\LVZERO}{{\lbl{L0-Typo}}\xspace}
\newcommand{\LVONE}{{\lbl{L1-Norm}}\xspace}
\newcommand{\LVTWO}{{\lbl{L2-Expl}}\xspace}
\newcommand{\LVTHREE}{{\lbl{L3-Coref}}\xspace}
\newcommand{\LVFOUR}{{\lbl{L4-Conn}}\xspace}
\newcommand{\LVMIX}{{\lbl{L5-Mix}}\xspace}
\newcommand{\LVFREE}{{\lbl{L6-Free}}\xspace}

\newcommand{\CLVZERO}{\textcolor{MidnightBlue}{\LVZERO}\xspace}
\newcommand{\CLVONE}{\textcolor{MidnightBlue}{\LVONE}\xspace} 
\newcommand{\CLVTWO}{\textcolor{MidnightBlue}{\LVTWO}\xspace} 
\newcommand{\CLVTHREE}{\textcolor{MidnightBlue}{\LVTHREE}\xspace} 
\newcommand{\CLVFOUR}{\textcolor{MidnightBlue}{\LVFOUR}\xspace}
\newcommand{\CLVMIX}{\textcolor{MidnightBlue}{\LVMIX}\xspace} 
\newcommand{\CLVFREE}{\textcolor{MidnightBlue}{\LVFREE}\xspace} 

\newcommand{\BParagraph}[1]{\noindent\textbf{#1}}



%% file: latex/0_abstract.tex
\begin{abstract}

Rewriting inputs to improve frozen downstream models has become a common strategy in modern NLP pipelines. Prior work on incremental dialogue discourse parsing (DDP) shows that supervised clarification models can rewrite fragmentary or underspecified utterances—such as resolving ellipsis or references—to improve parsing accuracy. In this work, we revisit this idea under realistic deployment conditions, where no clarification supervision is available and the clarifier must rely on zero-shot prompting or feedback from a frozen parser. Across three Segmented Discourse Representation Theory (SDRT) datasets and multiple parsers, we find that last-utterance clarification is far less reliable than suggested by supervised settings. Parser-agnostic rewriting often introduces more regressions than repairs, as edits that enable fixes also disrupt discourse cues relied upon by the parser. A best-of-8 rewriting analysis further reveals a practical ceiling: a large fraction of errors are not repairable through input rewriting alone. A parser-aware clarifier trained with GRPO reduces regressions by up to 37\% by learning conservative abstention, yet still fails to produce selectivity-aware clarifications that consistently improve parsing. Together, these findings recast clarification as a selective intervention problem. We identify \textbf{rewritability prediction}—deciding whether an utterance is repairable before intervention—as the key missing capability for input-side optimization of frozen discourse parsers, and a critical direction for improving agentic pipelines more broadly.\footnote{Data and code are available at \url{https://github.com/ounlp/Clarification-for-DDP}.} 
\end{abstract}

%% file: latex/1_intro.tex
\section{Introduction}
\begin{figure*}[t]
  \centering
  \includegraphics[width=0.98\textwidth]{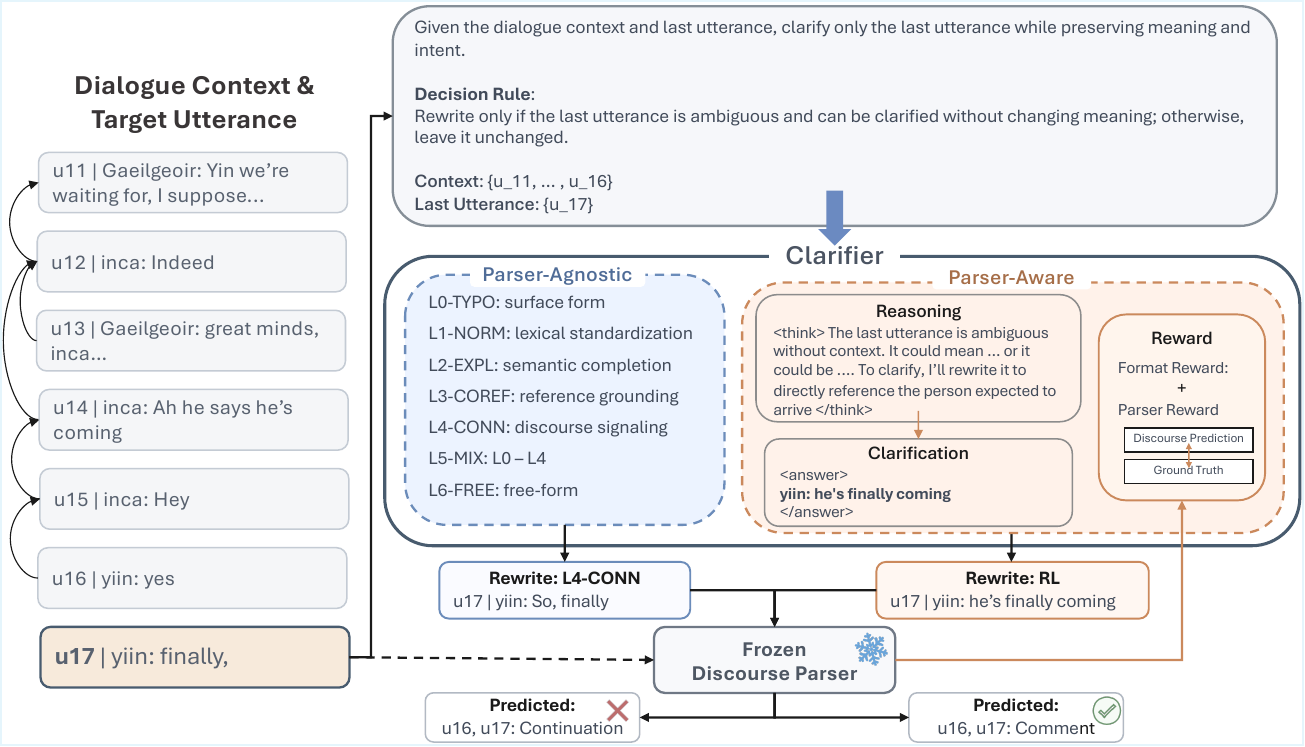}
  \caption{\textbf{Overview of our last-utterance clarification framework for incremental dialogue discourse parsing}. Given a dialogue context, the clarifier rewrites only the final utterance, replaces it in the parser’s context window, and passes the modified context to a frozen discourse parser. Parser-agnostic clarification uses predefined rewrite strategies, whereas parser-aware clarification optimizes the clarifier with reinforcement learning from parser-based reward computed by comparing the predicted discourse structure on the rewritten context against the ground truth.}
  \label{fig:running_example}
\end{figure*}

Recent work has shown that rewriting the input to a frozen system can improve NLP pipelines without retraining, including in machine translation~\citep{ki-carpuat-2025-automatic}, retrieval-augmented generation~\citep{wu-etal-2022-conqrr}, and conversational search~\citep{zhu-etal-2025-convsearch}. \citet{fan-etal-2025-improving} apply this idea to dialogue discourse parsing~(DDP) through \emph{last-utterance} clarification, training an LLM clarifier on supervised clarification data and then optimizing it with preference-based reinforcement learning~(RL) to rewrite the final utterance before parsing. However, such supervision is costly to obtain and difficult to reuse across parsers. We therefore ask whether last-utterance clarification can improve DDP without access to supervised clarification examples.


Unlike retrieval or translation, SDRT discourse parsing requires semantic inference over structured dialogue context~\citep{lascarides2007segmented}. An incremental parser such as Llamipa~\citep{thompson-etal-2024-llamipa} must determine how the current utterance connects to prior discourse. In a sliding-window setting, the last utterance becomes a natural intervention point—it is the only part of the current input that can still be rewritten before prediction. This makes last-utterance clarification particularly interesting, but also challenging: while clarification may reduce ambiguity, it can also alter surface cues (e.g., fragmentary syntax, pronouns, connectives) that the parser relies on.

Following~\citet{fan-etal-2025-improving}, we study this setting under incremental SDRT parsing, where no future context is available and past dialogue history and decisions cannot be revised. As illustrated in \autoref{fig:running_example}, the ambiguous final utterance (e.g., $u_{17}$: \emph{“finally,”}) is interpreted given the preceding context ($u_{11}$–$u_{16}$), and a clarifier must decide whether to intervene before passing it to a frozen parser. To go beyond supervised clarification, we consider two settings that reflect common agentic patterns:~(1) prompt-based interaction with a frozen tool or system (\emph{parser-agnostic}, \S\ref{ssec:parser-agnostic-strateiges}), and (2) adaptation through downstream feedback-driven learning to optimize performance~(\emph{parser-aware}, \S\ref{ssec:rl_rewriting}).

We first instantiate the parser-agnostic setting by evaluating seven clarification strategies, ranging from low-level typo correction to free-form semantic rewriting, without any access to parser feedback. In contrast, the parser-aware setting asks whether a clarifier can \emph{learn when and how to intervene} purely from downstream signals. To this end, we train a clarifier with GRPO~\citep{guo2025deepseek}, enabling it to acquire rewrite-or-copy behavior directly from frozen-parser feedback. Experiments are conducted on three SDRT datasets (STAC, Molweni, MSDC) and four parsers spanning both generative and discriminative families.

Our results reveal a consistent pattern. Parser-agnostic strategies often introduce more regressions than repairs, as edits that enable fixes can also disrupt cues the parser relies on. A parser-aware RL clarifier reduces regressions by up to 37\% by learning when to abstain, but still fails to produce clarifications that reliably improve parsing. A best-of-8 rewriting analysis further shows that about 80\% of errors are not repairable by any rewrite, pointing to~\textbf{rewritability prediction --- deciding when to intervene --- as a key direction forward}. Together, these findings suggest that last-utterance rewriting does not transfer cleanly to dialogue discourse parsing in an incremental setting, highlighting a limitation of common agentic designs for black-box components. More broadly, they indicate that upstream modules --- whether rewriting, planning, or tool orchestration --- are most effective not as generic prompt-driven interfaces, but when they are selective, context-aware, and aligned with downstream signals.

%% file: latex/2_related.tex
\section{Related Work}
\label{sec:related-work}

\paragraph{Dialogue Discourse Parsing.} 
DDP has been studied with both discriminative and generative parsers under multiple discourse formalisms, including Segmented Discourse Representation Theory~\citep[SDRT,][]{lascarides2007segmented,asher-etal-2016-discourse}, Rhetorical Structure Theory~\citep[RST,][]{mann1988rhetorical}, the Penn Discourse Treebank framework~\citep[PDTB,][]{prasad-etal-2008-penn}, and Dependency Dialogue Acts~\citep[DDA,][]{cai2023dependency,cai-etal-2025-search}. Early work focused on discriminative structure prediction for multi-party dialogue~\citep{afantenos-etal-2015-discourse,perret2016integer,shi2018deepsequentialmodeldiscourse,liu2021improving,chi-rudnicky-2022-structured}, while more recent work has shown that LLM-based generative parsers can achieve strong performance~\citep{li-etal-2024-dialogue,thompson-etal-2024-llamipa} on STAC~\citep{asher-etal-2016-discourse}, Molweni~\citep{li-etal-2020-molweni}, and MSDC~\citep{thompson-etal-2024-discourse}. Prior work has identified several input-side phenomena that are plausibly relevant to relation prediction, including unresolved references~\citep{choi-etal-2021-decontextualization}, fragmentary utterances~\citep{li-etal-2023-incomplete}, and explicit versus implicit discourse marking~\citep{liu-etal-2024-causes}. Our work treats these phenomena as potential targets of upstream clarification and tests whether resolving them actually helps a frozen parser.


\paragraph{Input Rewriting for Frozen Systems.} 
A growing body of work improves frozen downstream systems by rewriting their inputs at inference time, including in machine translation~\citep{sun-etal-2024-lcs, ki-carpuat-2025-automatic}, conversational retrieval and search~\citep{wu-etal-2022-conqrr,zhu-etal-2025-convsearch, cao-etal-2025-icr, xu2025rethinking}, retrieval-augmented generation~\citep{ma-etal-2023-query, ye-etal-2025-q} and recommendation system~\citep{lin2025rec}. This line of work is also related to broader meaning-preserving reformulation methods, such as decontextualization and incomplete utterance rewriting, which transform an input into a more explicit form without changing its underlying meaning~\citep{choi-etal-2021-decontextualization, li-etal-2023-incomplete}. Our setting differs from these tasks in that the downstream objective is not directly tied to local surface matching: in discourse parsing, rewriting a single utterance affects the parser only indirectly through its interaction with the broader dialogue context.


\paragraph{Clarification for Discourse Parsing.}
Most related to our work, \citet{fan-etal-2025-improving} 
train an LLM clarifier for DDP using supervised clarification data followed by preference-based reinforcement learning~\cite[RLHF;][]{ouyang2022training}. We study a different setting: the clarifier has access only to frozen parser feedback, with no supervised clarification data. This lets us test clarification as input-side intervention, rather than as supervised adaptation to parser-specific errors. \citet{aktas-roth-2025-clarifying} study connective insertion for underspecified discourse relations in instructional texts under PDTB, reporting gains from making implicit connectives explicit. Their work also frames clarification as a discourse-oriented rewriting operation, but in a different task setting. We therefore position our work as a study of the limits of clarification as input-side optimization for a frozen discourse parser: whether it remains effective without supervised clarification data, and whether frozen-parser feedback alone is sufficient to support useful intervention.


%% file: latex/4_methods.tex
\section{Methods and Experiments}
\label{sec:methods}

\subsection{Parser-Agnostic Clarification Strategies}
\label{ssec:parser-agnostic-strateiges}



Prior work has identified three input-side phenomena that specifically challenge discourse relation labeling~\citep{choi-etal-2021-decontextualization, ma-etal-2023-query, sun-etal-2023-improving, ki-carpuat-2025-automatic}. 
These observations motivate targeting these phenomena through upstream rewriting. However, whether resolving them actually helps a frozen parser, or whether the parser has learned to handle them through its training distribution, is precisely the empirical question we study. We therefore define clarification strategies at increasing levels of intervention, from conservative surface edits that are unlikely to disturb any parser cues, to deeper semantic edits that directly target the discourse-relevant phenomena above. This ordering lets us jointly ask: at what depth of intervention do repairs emerge, and at what depth do regressions begin to dominate?


\BParagraph{Surface form~(\CLVZERO)} corrects only obvious typos and surface errors, which may resolve entity mentions or cue words that affect attachment and relation labeling. Prior work on lexical normalization of non-canonical forms has been shown to improve downstream tagging performance, suggesting that restoring corrupted cue words can benefit structure prediction~\citep{van-der-goot-cetinoglu-2021-lexical}. 

\BParagraph{Lexical Standardization~(\CLVONE)} expands abbreviations/shorthand/slang into standard forms (e.g., “idk”→“I don’t know”), without altering wording or adding information. This reduces surface noise by restoring lexical cues that guide DDP, consistent with evidence that task-agnostic normalization improves robustness under noisy inputs \citep{bitton-etal-2022-adversarial}.

\BParagraph{Semantic Completion~(\CLVTWO)} targets ellipsis and fragmentary turns by minimally restoring omitted but context-entailed material, making the last-utterance more self-contained. \citet{li-etal-2023-incomplete} show that rewriting underspecified dialogue utterances supports the benefit of meaning-preserving completion for dialogue understanding.

\BParagraph{Reference Grounding~(\CLVTHREE)} makes coreference explicit by replacing pronouns/deictics with contextually unambiguous antecedents to improve entity continuity and reduce referential ambiguity~\citep{choi-etal-2021-decontextualization}. This aligns with evidence from conversational QA/retrieval that making context-dependent turns more explicit can improve downstream performance \citep{wu-etal-2022-conqrr}.

\BParagraph{Discourse Signaling~(\CLVFOUR)} adds a semantically compatible cue (e.g, “because,” “but,” “so,” “by the way”) to signal the intended discourse relation explicitly. Such connectives often constrain relation interpretation, and their removal may trigger label shift \citep{liu-etal-2024-causes}. Therefore, we treat connective insertion as a natural parser-agnostic clarification strategy, following~\citet{aktas-roth-2025-clarifying}.

 In addition to the five base levels~(\textcolor{MidnightBlue}{L0 to L4}), we include a mixed-rule strategy~(\CLVMIX) and a minimally constrained free-form clarification strategy~(\CLVFREE), yielding 7 parser-agnostic clarifiers. All strategies share the same prompt skeleton and differ only in their level-specific rules. \CLVMIX exposes all five base-level edit operations~(\textcolor{MidnightBlue}{L0 to L4}) within a single prompt and permits the clarifier to apply any subset whose per-rule safety conditions are satisfied. As shown in \autoref{tab:mb-bertscore}, for each level, we select the prompt that ensures clarification while maximizing meaning preservation metric BERTScore~\citep{zhang2019bertscore} on STAC validation set under that level's corresponding edit constraints~\footnote{Our prompting experiments suggest that stronger constraints are necessary to elicit repairs. Please refer to Appendix~\ref{sec:appendix-prompts}.}.




\subsection{RL-based Parser-Aware Clarification}
\label{ssec:rl_rewriting}
We train a rewriting policy $\pi_{\phi}$ with reinforcement learning to improve a frozen discourse parser. At each dialogue step $t$, the policy observes the current dialogue prefix $s_t = (u_1, \ldots, u_t),$ where $u_t$ is the final utterance to be clarified, and generates either the original utterance or a rewritten version $\tilde{u}_t$. We then replace $u_t$ with $\tilde{u}_t$ in the dialogue prefix and run the frozen parser on both the original and rewritten contexts. The parser's predictions are compared against the gold discourse structure, and the resulting change in parsing quality is converted into a scalar reward for policy learning.

We optimize $\pi_\phi$ with GRPO~\citep{guo2025deepseek}, using a clipped policy-gradient objective with group-relative advantages and KL regularization to prevent degenerate rewrites. We use a verifiable reward defined entirely by the frozen parser's behavior on the rewritten utterance. The reward has two components: a format reward and a parsing reward. Following~\citet{guo2025deepseek} and \citet{jiang2025deepretrieval}, who demonstrate that enforcing structured output via format rewards stabilizes GRPO training and prevents degenerate generations, we assign $r_{fmt}=1$ when the output contains exactly one valid \texttt{<think>…<answer>} block in order, $r_{fmt}=-2$ otherwise. For parsing rewards, we assign $+2$ for correcting errors, $-2$ for introducing errors, $+1$ for partial fixes, and $0$ otherwise. This shaping encourages well-formed clarifications that measurably improve discourse parsing. Detailed RL implementation using verl~\citep{sheng2024hybridflow} is provided in \S\autoref{sec:appendix-rl}.


\subsection{Experiment Setup}
\label{ssec:exp}

\paragraph{Datasets.}
We evaluate on three SDRT-based dialogue discourse parsing datasets: STAC, Molweni, and MSDC, using their standard train/dev/test splits~(Table~\ref{tab:dataset}). STAC contains multi-party game dialogues, Molweni includes Ubuntu technical chats, and MSDC consists of collaborative Minecraft dialogues. Importantly, we do not use any clarification or rewriting supervision—our methods operate without clarification data.



\begin{table}[t]
\centering
\small
\begin{tabular}{lccc}
\toprule
\textbf{Dataset} & \textbf{Train} & \textbf{Dev} & \textbf{Test} \\
\midrule
STAC    & 9,507 & 1,084 & 1,045 \\
Molweni & 9,215 & 1,026 & 1,107 \\
MSDC    & 1,732 & 1,016 & 989 \\
\bottomrule
\end{tabular}
\caption{Dataset statistics for SDRT dialogue discourse parsing datasets: STAC, Molweni, and MSDC. Counts are the number of last-utterance parsing instances (one per dialogue turn) in each split. We follow the standard splits provided with each dataset.}
\label{tab:dataset}
\end{table}

\paragraph{Models.} On each dataset, we train two LLM-based generative parsers by supervised LoRA fine-tuning~\citep{hu2021lora} from \emph{Qwen3-8B} and \emph{Qwen3-14B}~\citep{yang2025qwen3technicalreport}, outperforming the original Llama3-based Llamipa models~(as \autoref{tab:llm-parser-results}, additional training details are provided in \S\ref{ssec:sft-parser}). To test whether our findings generalize beyond generative parsers, we also evaluate SDDP~\citep{chi-rudnicky-2022-structured}, a discriminative graph-based discourse parser that uses a BERT-based encoder to score candidate parent-child links and relation types via biaffine attention~\citep{chen2014fast}, then selects a globally consistent set of links. We adapt SDDP to our incremental setting by parsing each dialogue prefix and evaluating only the relations predicted for the current utterance.

We study seven parser-agnostic~(\S\ref{ssec:parser-agnostic-strateiges}) and one parser-aware~(\S\ref{ssec:rl_rewriting}) clarifiers based on Qwen3 and Qwen2.5-7B-Instruct respectively.\footnote{We use Qwen2.5-7B-Instruct rather than the Qwen3 for the parser-aware clarifier due to compatibility constraints with our RL training framework~(verl).} We train the parser-aware clarifier against the Qwen3-8B parser and report its results in comparison with the \CLVFREE baseline on the same parser. Because the clarifier and parser do not share parameters and interact only through rewritten text, this model mismatch does not affect the validity of our setup.

\paragraph{Evaluation.} We use micro-F1 for link attachment and full structure (link+relation) to measure model performance across all datasets, parsers, and clarifiers. We also report the number of fixes (wrong$\rightarrow$correct; ``fix'') and regressions (correct$\rightarrow$wrong; ``reg'') after clarifying the last utterance (\autoref{tab:llm-parser-results}). 


%% file: latex/5_analysis.tex
\section{Main Results and Analysis}
\label{sec:analysis}

Our results show a clear pattern: last-utterance clarification often harms more than it helps, and even with RL, the challenge is not how to rewrite, but when to intervene. Parser-agnostic strategies~(\S\ref{ssec:rq1-llm-parser}) introduce more regressions than repairs, while parser-aware RL~(\S\ref{ssec:rq3-rl-parser-aware}) learns dataset-level intervention regimes without achieving reliable per-instance selectivity. Finally, we provide further insights via best-effort baseline analysis and ablation studies~(\S\ref{sec:discussion}).


\begin{table}[t]
\centering
\scriptsize
\setlength{\tabcolsep}{4pt} 
\begin{tabular}{llccccc}
\toprule
\multirow{2}{*}{\textbf{Method}} &
\multicolumn{2}{c}{\textbf{STAC}} &
\multicolumn{2}{c}{\textbf{Molweni}} &
\multicolumn{2}{c}{\textbf{MSDC}} \\
\cmidrule(lr){2-3}\cmidrule(lr){4-5}\cmidrule(lr){6-7}
& \textbf{F1} & \textbf{Fix/Reg} & \textbf{F1} & \textbf{Fix/Reg} & \textbf{F1} & \textbf{Fix/Reg} \\
\hline
Llamipa & 0.577 & - & - & - & 0.795 & - \\ 
\hline
\addlinespace[3pt]
Qwen3-8B & 0.598 & - & 0.571 & - & 0.800 & - \\
\CLVZERO & \textbf{0.596} & \textbf{2/3} & \underline{0.565} & \underline{28/51} & \textbf{0.800} & \textbf{11/11} \\

\CLVONE& 0.594 & 5/9 & 0.564 & 31/59 & \underline{0.797} & \underline{12/24} \\
\CLVTWO & 0.582 & 7/23 & \textbf{0.568} & \textbf{33/47} & 0.795 & 24/48 \\
\CLVTHREE & \underline{0.594} & \underline{5/8} & 0.556 & 38/96 & 0.794 & 14/40 \\
\CLVFOUR  & 0.540 & 21/88 & 0.531 & 74/234 & 0.771 & 61/212 \\
\CLVMIX  & 0.589 & 9/20 & 0.555 & 50/114 & 0.793 & 25/56 \\
\CLVFREE  & 0.577 & 17/38 & 0.565 & 104/128 & 0.789 & 43/105 \\
\hline

\addlinespace[3pt]
Qwen3-14B & 0.591 & - & 0.554 & - & 0.805 & - \\
\CLVZERO  & \textbf{0.589} & \textbf{2/4} & \underline{0.552} & \underline{17/27} & 0.803 & 2/10 \\
\CLVONE & 0.585 & 6/12 & \textbf{0.552} & \textbf{34/42} & 0.804 & 9/12 \\
\CLVTWO  & 0.576 & 10/26 & 0.551 & 36/51 & \textbf{0.799} & \textbf{21/47} \\
\CLVTHREE  & \underline{0.587} & \underline{3/7} & 0.538 & 30/94 & 0.799 & 12/45 \\
\CLVFOUR  & 0.532 & 22/89 & 0.512 & 73/239 & 0.779 & 60/192 \\
\CLVMIX  & 0.584 & 9/17 & 0.549 & 49/74 & \underline{0.799} & \underline{13/44} \\
\CLVFREE  & 0.567 & 16/44 & 0.549 & 90/111 & 0.797 & 40/84 \\
\bottomrule
\end{tabular}
\caption{Parser-agnostic clarification introduces more regressions than repairs, with discourse connective insertion (\CLVFOUR) producing the largest degradations.}
\label{tab:llm-parser-results}
\end{table}

\subsection{Parser-Agnostic Rewriting Induces Regressions}
\label{ssec:rq1-llm-parser}


Table~\ref{tab:llm-parser-results} summarizes the impact of seven parser-agnostic clarification strategies applied to the last utterance under two frozen parsers (Qwen3-8B/14B) across three datasets. A consistent pattern emerges: \textit{parser-agnostic clarification introduces more regressions than repairs, indicating that improved surface clarity does not reliably benefit frozen discourse parsing.} For Qwen3-8B parser, all strategies either leave performance nearly unchanged~(e.g., STAC \CLVZERO: 0.598 $\rightarrow$ 0.596) or reduce Link+Rel F1, with the largest drops consistently observed for discourse connective insertion~(\CLVFOUR). Qwen3-14B shows the same trend, including the same failure mode for \CLVFOUR, suggesting that the limitation is not specific to smaller models.

A second pattern reveals a clear safety–utility trade-off across intervention depth. Conservative strategies~(\CLVZERO and \CLVONE) are relatively safe but offer limited gains (e.g., on STAC with Qwen3-8B: 2 vs.~3 and 5 vs.~9 for repairs vs. regressions). In contrast, more semantic interventions create greater repair opportunities but incur substantially more regressions: \CLVFOUR yields 21 vs. 88, and \CLVFREE 17 vs. 38. This imbalance persists across models and datasets, with heavier interventions showing greater instability. \textit{In short, the same edits that enable fixes also disproportionately increase the risk of harming correct predictions.}

\begin{figure*}[!t]
\centering

\begin{subfigure}[t]{0.32\textwidth}
    \centering
    \includegraphics[width=\linewidth]{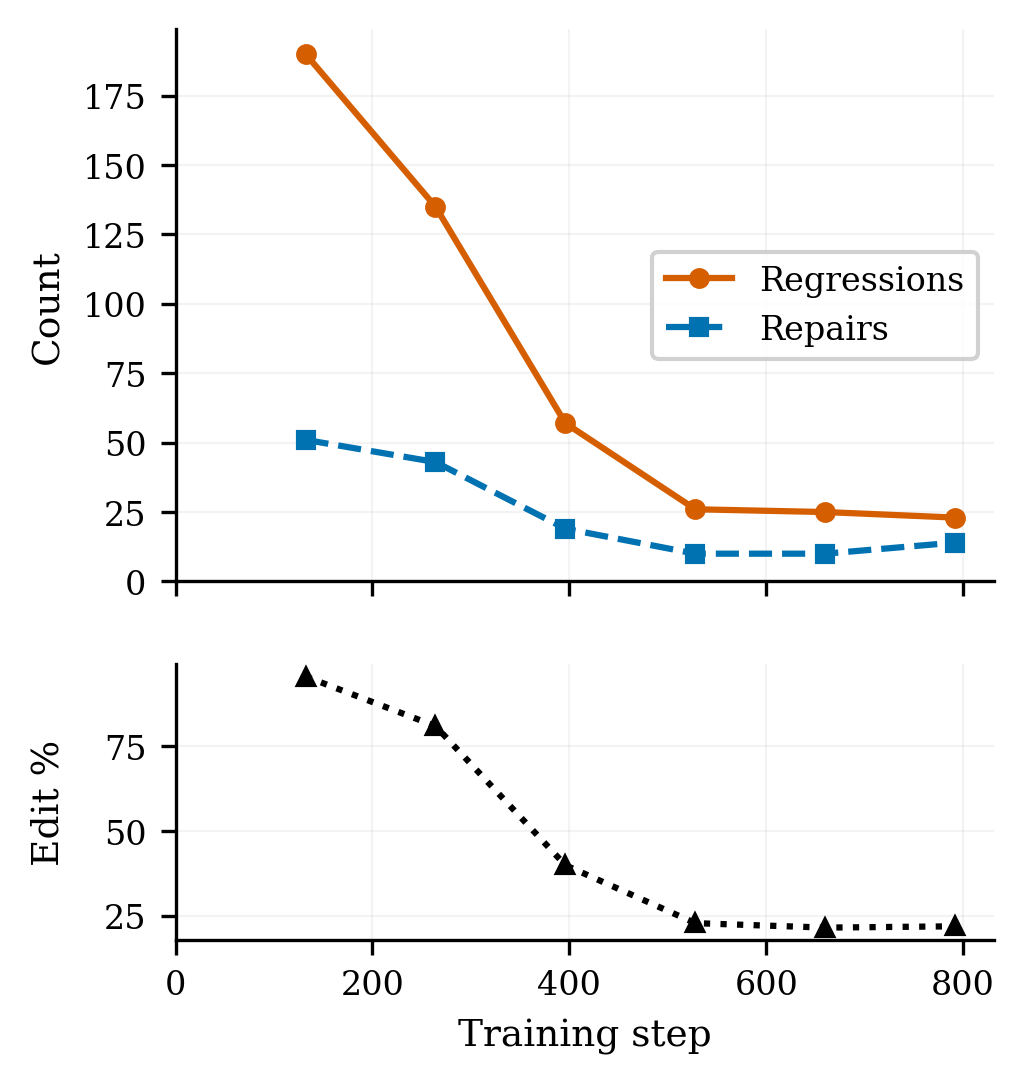}
    \caption{STAC}
\end{subfigure}
\hfill
\begin{subfigure}[t]{0.32\textwidth}
    \centering
    \includegraphics[width=\linewidth]{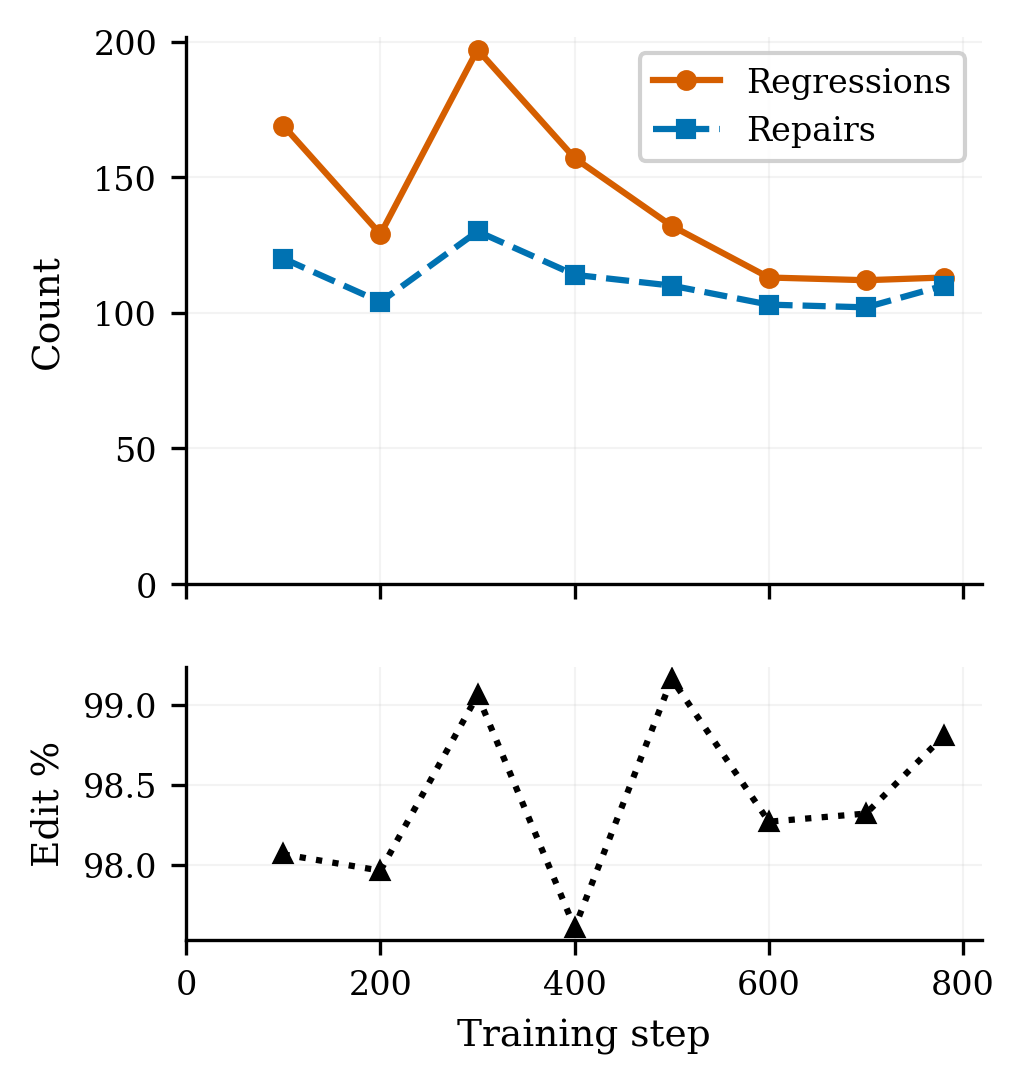}
    \caption{Molweni}
\end{subfigure}
\hfill
\begin{subfigure}[t]{0.32\textwidth}
    \centering
    \includegraphics[width=\linewidth]{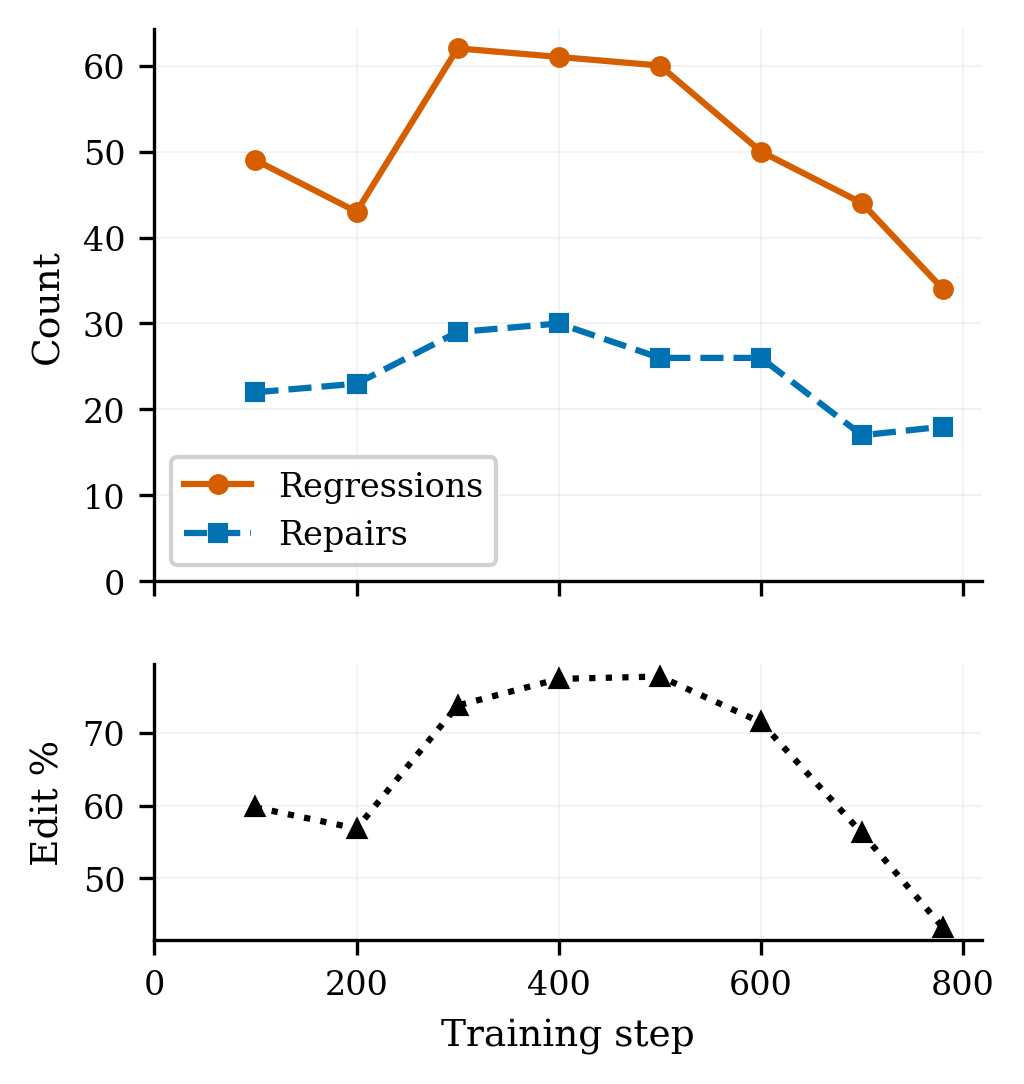}
    \caption{MSDC}
\end{subfigure}

\caption{Validation-set learning dynamics of the parser-aware clarifier on STAC, Molweni, and MSDC. Across datasets, GRPO training produces different intervention behaviors rather than a single stable selective policy: conservative abstention on STAC, near-universal rewriting on Molweni, and unstable behavior on MSDC.}
\label{fig:training_curves}
\end{figure*}

\begin{table}[t]
\centering
\small
\setlength{\tabcolsep}{4pt}
\begin{tabular}{lccc}
\toprule
\textbf{Metric} & \textbf{STAC} & \textbf{Molweni} & \textbf{MSDC} \\
\midrule
\multicolumn{4}{l}{\textit{LLM-rewrite (\CLVFREE)}} \\
\midrule
Link         & 0.752 & -    & -    \\
Link+Rel          & 0.577 & 0.565 & 0.789 \\
Repairs      & 17    & 104   & 43    \\
Regressions  & 38    & 128   & 105   \\
Edit\%       & 0.499 & 0.882 & 0.453 \\
\midrule
\multicolumn{4}{l}{\textit{Parser-aware RL}} \\
\midrule
Link         & 0.754 & 0.846    & 0.872    \\
Link+Rel          & 0.586 & 0.551    & 0.784    \\
Repairs      & 7     & 110   & 51    \\
Regressions  & 24    & 113   & 125   \\
Edit\%       & 0.214 & 0.988 & 0.338 \\
\bottomrule
\end{tabular}
\caption{Results of parser-aware clarification compared with the L6-FREE baseline across three datasets using the Qwen3-8B parser. Edit\% is the fraction of examples where the last utterance is rewritten. The learned policy behaves differently across datasets: it becomes more conservative on STAC, rewrites nearly all utterances on Molweni, and remains harmful on MSDC.}
\vspace{-3pt}
\label{tab:parser_aware_rl}
\end{table}

\subsection{Parser-Aware RL Learns Intervention Regimes, Not Selective Policies}
\label{ssec:rq3-rl-parser-aware}

The key question for parser-aware clarification is whether frozen-parser feedback can train a policy that intervenes selectively—rewriting when helpful and abstaining when harmful. Table~\ref{tab:parser_aware_rl} and Figure~\ref{fig:training_curves} show that RL does learn coherent, convergent behaviors within each dataset. However, rather than collapsing to a single universal strategy, the learned policies diverge in systematic ways: conservative abstention on STAC, near-universal rewriting on Molweni, and a more unstable mixed behavior on MSDC. This variation suggests that the learning signal shapes how often to intervene, but not yet when to intervene at the instance level.

On STAC, the learned policy improves Link+Rel F1 over the \CLVFREE baseline while substantially reducing regressions (38 $\rightarrow$ 24, -37\%). Figure~\ref{fig:training_curves} shows a corresponding drop in Edit\% from 49.9\% to 21.4\%, indicating that the policy learns to abstain. Notably, this gain comes not from better rewrites --- repairs decrease (17 to 7) --- but from avoiding harmful ones. RL therefore discovers a conservative regime where doing less is better.

This abstention dynamic does not generalize to the other two datasets. On Molweni, the learned policy rewrites nearly every utterance, reaching a final Edit\% of 98.8\%. As shown in Table~\ref{tab:parser_aware_rl}, regressions fall only modestly (128 $\rightarrow$ 113) and repairs increase only slightly (104 $\rightarrow$ 110), while Link+Rel F1 drops relative to the~\CLVFREE baseline. The policy therefore learns to intervene almost universally rather than selectively. MSDC exhibits a different behavior. Figure~\ref{fig:training_curves} shows that Edit\% eventually declines during training, but regressions remain high and the final policy is still harmful: compared with \CLVFREE, regressions increase (105 $\rightarrow$ 125) while repairs rise only modestly (43 $\rightarrow$ 51), resulting in a lower Link+Rel F1. Unlike STAC, reducing the intervention rate is not sufficient here to avoid harmful rewrites.

Taken together, although the learned policies vary across datasets, they converge to structured intervention regimes rather than random behavior, yet still fall short of true per-instance selectivity. 

%% file: latex/6_discussion.tex
\section{Ablation Studies and Discussion}
\label{sec:discussion}

The main results suggest that the central challenge is not whether clarification can help, but when and why it does. To better understand this, we conduct further analyses to characterize the repairability ceiling of last-utterance rewriting~(\S\ref{ssec:oracle}), examine whether the observed instability depends on parser family~(\S\ref{ssec:rq2-diff-parsers}), and synthesize key implications for future work~(\S\ref{ssec:failure-analysis}).


\subsection{How Repairable Are Parser Errors?}
\label{ssec:oracle}


To estimate the headroom of clarification on DDP, we construct a best-effort baseline by generating eight independent rewrites on the validation set with the base Qwen3-8B model(our skeleton model for RL-post-training). For each incorrect prediction with the original last utterance, we use the same~\CLVFREE strategy to sample 8 free-form rewrites of the last utterance, rerun the frozen parser, and mark the example as repairable if any rewrite yields the correct prediction. This estimates the upper bound of recoverable errors under repeated rewriting, rather than the performance of a practical clarifier.

\begin{table}[!t]
\centering
\resizebox{\columnwidth}{!}{
\begin{tabular}{l c cc cc cc}
\toprule
& & \multicolumn{2}{c}{\textbf{STAC}} & \multicolumn{2}{c}{\textbf{Molweni}} & \multicolumn{2}{c}{\textbf{MSDC}} \\
\cmidrule(lr){3-4} \cmidrule(lr){5-6} \cmidrule(lr){7-8}
\textbf{Category} & \textbf{Succ. (of 8)} & \textbf{\# Ex.} & \textbf{\%} & \textbf{\# Ex.} & \textbf{\%} & \textbf{\# Ex.} & \textbf{\%} \\
\midrule
Not repairable        & 0    & 475 & 80.0 & 1173 & 72.5 & 265 & 71.8 \\
Rarely repairable     & 1--2 & 66  & 11.1 & 182  & 11.2 & 35  & 9.5  \\
Moderately repairable & 3--5 & 39  & 6.6  & 147  & 9.1  & 52  & 14.1 \\
Highly repairable     & 6--8 & 14  & 2.4  & 116  & 7.2  & 17  & 4.6  \\
\bottomrule
\end{tabular}
}
\caption{Distribution of originally incorrect examples on validation set by estimated repairability under a best-of-8 free-form rewriting baseline. Repairability is defined by how many of eight sampled rewrites, generated from the same free-form prompt, yield a correct prediction.}
\label{tab:repairability}
\end{table}

\begin{table}[!t]
\centering
\small
\begin{tabular}{lcccc}
\toprule
\textbf{Dataset} & \textbf{Not rep.} & \textbf{Rarely} & \textbf{Moderately} & \textbf{Highly} \\
\midrule
STAC    & 0.9  & 4.4 & \textbf{12.5} & 11.1 \\
Molweni & 0.8  & 7.7 & 19.7 & \textbf{31.9} \\
MSDC    & 1.5  & 5.7 & 9.6 & \textbf{23.5} \\
\bottomrule
\end{tabular}
\caption{Repair rate (\%) of the parser-aware RL clarifier on baseline-wrong validation examples, grouped by estimated repairability from Table~\ref{tab:repairability}. For each bucket, the repair rate is the fraction of examples for which the final RL policy changes an originally incorrect prediction into a correct one.}
\label{tab:rl_by_repairability}
\end{table}

Table~\ref{tab:repairability} shows that the repairability ceiling is low across all three datasets. Most parser errors remain unrepaired even under candidate selection: $80.0\%$ errors on STAC, $72.5\%$ on Molweni, and $71.8\%$ on MSDC are not repaired by any of the eight sampled rewrites. At the same time, repairability appears as a spectrum rather than a binary property. A non-trivial minority of cases are moderately repairable, accounting for $6.6\%$ of errors on STAC, $9.1\%$ on Molweni, and $14.1\%$ on MSDC, while only a much smaller fraction are highly repairable. These results indicate that repeated free-form rewriting of the final utterance offers genuine headroom for a limited subset of cases, but leaves most errors untouched. 


To examine whether the parser-aware RL policy aligns with the repairable subset -- and potentially goes beyond it -- we group baseline-wrong validation examples by their estimated repairability~(Table~\ref{tab:repairability}) and compute RL repair rates within each bucket. Table~\ref{tab:rl_by_repairability} shows a clear trend: the policy rarely repairs examples deemed not repairable by the candidate rewrites (<2\% across datasets), while repair rates increase substantially with estimated difficulty. On Molweni and MSDC, this rise is nearly monotonic, reaching 31.9\% and 23.5\% in the highly repairable bucket; STAC shows a similar pattern, with minor variance due to small sample size. Notably, even within the ``not-repairable'' bucket, the small fraction of successful repairs (<2\%) highlights the exploratory power of RL: the policy can occasionally discover effective rewrites beyond the finite candidate rewrites. \textbf{Overall, candidate rewriting repairability is a strong predictor of RL success, while RL retains a limited but meaningful ability to search beyond it}. However, this alignment reflects the quality of rewrites, not the decision to rewrite: on STAC, RL's rewrite rate stays near the 23.0\% base rate across all four buckets, confirming global abstention rather than per-instance selectivity.


Tables~\ref{tab:repairability} and~\ref{tab:rl_by_repairability} do more than explain why overall gains are limited—they reveal a structured landscape of opportunities. While a large portion of cases appears inherently unrepairable under local rewriting, a meaningful subset remains consistently repairable, and RL shows a clear ability to align with --- even occasionally go beyond --- this structure through exploration. This perspective reframes the problem: rather than treating clarification as uniformly applicable, it suggests a more selective paradigm. A natural next step is rewritability prediction (identifying when intervention is likely to help),  so that future systems can focus effort where improvement is possible, rather than attempting to rewrite indiscriminately.



\subsection{Do Different Parser Families Matter?}
\label{ssec:rq2-diff-parsers}


\begin{table}[!t]
\centering
\small
\begin{tabular}{lcccc}
\toprule
\textbf{Method} & \textbf{Link} & \textbf{Link + Rel} & \textbf{Fix/Reg} & \textbf{Edit\%} \\
\midrule
\multicolumn{5}{c}{\textbf{STAC}} \\
\midrule
SDDP              & 0.729 & 0.557 & --    & --    \\
\CLVZERO  & 0.727 & \textbf{0.552} & \textbf{0/4}   & 0.098 \\
\CLVONE  & 0.727 & \underline{0.551} & \underline{6/10}  & 0.207 \\
\CLVTWO  & 0.726 & 0.539 & 9/27  & 0.216 \\
\CLVTHREE & 0.725 & 0.549 & 3/13  & 0.083 \\
\CLVFOUR  & 0.706 & 0.497 & 19/81 & 0.462 \\
\CLVMIX   & 0.721 & 0.547 & 10/22 & 0.282 \\
\CLVFREE  & 0.717 & 0.525 & 17/47 & 0.499 \\
\midrule
\multicolumn{5}{c}{\textbf{Molweni}} \\
\midrule
SDDP              & 0.798 & 0.543 & --     & --    \\
\CLVZERO  & 0.795 & \textbf{0.537} & \textbf{16/40}  & 0.425 \\
\CLVONE  & 0.797 & \underline{0.536} & \underline{18/46}  & 0.448 \\
\CLVTWO  & 0.799 & 0.535 & 22/54  & 0.244 \\
\CLVTHREE & 0.796 & 0.529 & 22/75  & 0.297 \\
\CLVFOUR  & 0.797 & 0.522 & 76/158 & 0.684 \\
\CLVMIX   & 0.793 & 0.530 & 45/96  & 0.691 \\
\CLVFREE  & 0.794 & 0.533 & 74/111 & 0.882 \\
\midrule
\multicolumn{5}{c}{\textbf{MSDC}} \\
\midrule
SDDP  & 0.779 & 0.696 & --     & --    \\
\CLVZERO  & 0.778 & \textbf{0.696} & \textbf{11/18} & 0.070 \\
\CLVONE  & 0.778 & \underline{0.696} & \underline{13/20} & 0.123 \\
\CLVTWO  & 0.776 & 0.695 & 8/24 & 0.147 \\
\CLVTHREE & 0.776 & 0.694 & 19/37  & 0.123 \\
\CLVFOUR  & 0.768 & 0.688 & 20/79 & 0.458 \\
\CLVMIX   & 0.775 & 0.694 & 12/32  & 0.196 \\
\CLVFREE  & 0.775 & 0.693 & 19/43 & 0.453 \\
\bottomrule
\end{tabular}
\caption{Parser-agnostic clarification results with the SDDP parser on STAC, Molweni, and MSDC. The same pattern holds across all three corpora: conservative edits are near-neutral, while discourse connective insertion (\CLVFOUR) produces the largest degradation.}
\vspace{-4pt}
\label{tab:sddp_results}
\end{table}

We evaluate SDDP on STAC, Molweni and MSDC~(Table~\ref{tab:sddp_results}) and compare it with the LLM-based parsers~(Table~\ref{tab:llm-parser-results}), and we observe that the same clarification strategies yield qualitatively similar patterns across parser families. Conservative surface-level edits (\CLVZERO, \CLVONE) are largely neutral across both LLM-based and SDDP parsers, producing small changes in Link+Rel with relatively few repairs or regressions. In contrast, discourse connective insertion (\CLVFOUR) produces the largest degradations on both families.

Notably, structured decoding does not eliminate clarification-induced regressions. Despite SDDP's explicit global structure modeling, it exhibits the same overall pattern as the LLM-based parsers: conservative edits are mostly neutral, while semantically heavier rewrites reduce Link+Rel accuracy. Strategies with higher intervention rates, especially \CLVFOUR, \CLVMIX, and \CLVFREE, are again associated with larger performance drops. These results show that the instability of parser-agnostic clarification is not specific to LLM-based parsers, but appears across parser families. On MSDC, SDDP exhibits the same qualitative pattern: conservative edits (\CLVZERO, \CLVONE) remain near-neutral, while \CLVFOUR causes the largest degradation (20/79 fixes vs. regressions), consistent with the LLM parsers on the same dataset.

One caveat: since choosing an appropriate connective is itself a hard generation task~\citep{liu-etal-2024-causes}, \CLVFOUR's regressions may partly reflect poor marker choice rather than distribution shift. We read distribution shift as the main driver, since the easy-to-execute conservative strategies (\CLVZERO, \CLVONE) still edit non-trivially yet stay near-neutral, and leave fully disentangling the two to future work.

\begin{table*}[!t]
\centering
\scriptsize
\setlength{\tabcolsep}{3pt}
\begin{tabular}{clp{5.0cm}p{5.0cm}p{3.0cm}}
\toprule
& \textbf{Original} & \textbf{RL rewrite} & \textbf{Candidate rewrites} & \textbf{Gold} \\
\midrule
\textbf{A}
  & really
  & \textbf{oh, you mean 7 again?}
  & \{``really''\}
  & \textsc{Clarification\_question} \\
\textbf{B}
  & Sorry, sheep
  & \textbf{Sorry, I meant sheep}
  & \{``But, Sorry, sheep'', ``Sorry, ore'', ``Sorry, sheep''\}
  & \textsc{Correction} \\
\textbf{C}
  & I have no sheep :)
  & \textbf{I don't have any sheep but I'm offering wheat :)}
  & \{``But I have no sheep :)'', ``I have no sheep :)''\}
  & \textsc{Elaboration} \\
\textbf{D}
  & tough times..
  & \textbf{I'm out of resources.}
  & \{``So tough times..''\}
  & \textsc{Explanation} \\
\bottomrule
\end{tabular}
\caption{RL-generated rewrites that lie outside the LLM rewrite candidate pool yet improve parser output. The ``Candidate rewrites'' column lists only representative examples from the rewrites produced by the template strategies~(\CLVTWO--\CLVFREE); additional candidates may exist but are omitted for space. The RL rewrite is absent from the complete pool for every utterance.
}
\label{tab:rl-beyond-oracle}
\end{table*}

\subsection{What Works and What Remains Challenging?}
\label{ssec:failure-analysis}

First, clarification is effective on a well-defined subset of cases. The repairability analysis (\S\ref{ssec:oracle}) shows that a non-trivial portion of errors are consistently recoverable through last-utterance rewriting, establishing genuine headroom for improvement. The parser-aware RL results further demonstrate that \textbf{learning-based approaches can align with this structure}: RL increases repair rates on more repairable instances and even exhibits exploratory capability, occasionally discovering successful rewrites beyond the finite rewriting candidates. In addition, across parser families, conservative interventions (\CLVZERO, \CLVONE) provide stable behavior with minimal regressions, indicating that \textbf{safe intervention regimes are achievable}.

Meanwhile, clarification is inherently selective. A large portion of errors might lie outside the reach of local rewriting, and more aggressive interventions --- while enabling additional repairs --- introduce more regression risk. This trade-off appears consistently across both LLM-based and structured parsers, suggesting it reflects a general property of discourse parsing rather than a model-specific limitation. In the parser-aware setting, RL can influence global intervention behavior (e.g., learning to abstain), but does not yet yield a stable, fine-grained per-instance decision policy across datasets. Together, these results indicate that the main challenge is not only generating better rewrites, but \textbf{reliably identifying when rewriting is beneficial}.

These findings point to several promising directions. First, the structured repairability landscape motivates \textbf{rewritability prediction} as a key next step: estimating whether an instance is locally repairable before applying any intervention. Second, the presence of unrecoverable cases suggests moving beyond single-utterance rewriting \textbf{toward broader, context-level interventions} that can address dependencies outside the local window. 
More broadly, our results refine a common agentic-AI pattern: when intervention is limited to local, prompt-based rewriting over a fixed interface, gains are constrained by both repairability and the difficulty of selective application. This suggests that effective pipelines require \textbf{selective intervention, broader control over context, and better alignment with downstream signals}. \textit{This extends beyond rewriting: upstream components — planning, routing, tool orchestration — are unlikely to be reliably effective as generic, prompt-driven front-ends; their success depends on being selective, context-aware, and tightly coupled with downstream structure and feedback.}

\subsection{Qualitative Analysis}
\label{sec:qualitative}

\paragraph{SFT parser sensitivity to surface form.}
Fixing a typo or expanding an abbreviation can flip the parser's discourse prediction, even when the meaning is unchanged to human (see Table~\ref{tab:l0l1-surface} in the appendix).
On the fix side, normalizing \textit{yeah} to \textit{yes} changes the parser's prediction from Acknowledgement to the correct Question-answer\_pair; expanding the abbreviation \textit{go nz} to \textit{go New Zealand} shifts a mispredicted Continuation into the correct Result. On the regression side, expanding \textit{lol} to \textit{laughing out loud} breaks a correct Comment prediction to Elaboration; capitalizing \textit{peurto rico} to \textit{Puerto Rico}---a change no human would register---causes a correct Contrast to be mispredicted as Comment. These cases show the SFT parser has not fully abstracted away lexical surface form, which is part of what motivates the rewriting pipeline and the shift to RL training on parser reward.

\paragraph{When not to rewrite, and when rewriting is not enough.}
The RL agent also learns when to abstain. Bare fragments like \textit{``what for?''} and non-verbal tokens like \textit{``\textasciicircum\textasciicircum''}
are already parsed correctly in context; the agent leaves them unchanged, avoiding the noise that would come from always appending a paraphrase. At the same time, some errors lie beyond the reach of any surface rewrite. When \textit{``if you roll a 7''} is completed to
\textit{``if you roll a 7, you can move him''}---a factually correct
answer---both the original and the rewrite still attach to the wrong
antecedent question. The mistake is in discourse antecedent selection across two consecutive questions in a multi-party context, a structural ambiguity that rephrasing the answer cannot resolve.

\paragraph{RL discovering rewrites beyond the LLM rewriting candidates.}
Table~\ref{tab:rl-beyond-oracle} gives four cases where the RL rewrite is absent from the LLM rewrite candidate pool (\S\ref{ssec:oracle}, \S\ref{ssec:failure-analysis}). Rule strategies produce only surface variants within their rules; RL instead, guided by parser reward, escapes these constraints in two distinct ways.

When the candidate rewrites cannot effectively recover the context, RL recovers the correct discourse function from scratch: recognizing that \textit{``really''} functions as a Clarification\_question and rewriting explicitly with a question mark accordingly (\textit{``oh, you mean 7 again?''}). In row B, RL instead finds the minimal precise repair---inserting \textit{``I meant''}---to mark the self-correction explicitly as a Correction (\textit{``Sorry, I meant sheep''}).
In row C the candidates can only prepend \textit{``But''}; RL reads the preceding turn (\textit{``I'm definitely giving wheat''}) and incorporates it explicitly, recovering an Elaboration relation the candidate rewrites cannot reach. And in row D it can only prepend a connective (leaving the utterance idiomatic), yet RL translates the colloquial expression into an explicit proposition (\textit{``tough times..''} $\to$ \textit{``I'm out of resources.''}), recovering an Explanation relation the candidate rewriting strategies cannot reach. Together, these cases suggest RL acquires a sense of discourse function that template strategies, by design, cannot express.


%% file: latex/7_conclusion.tex
\section{Conclusion and Future Work}
\label{sec:conclusion}



We examine a simple question: can making dialogue utterances clearer reliably improve a frozen discourse parser? Our results suggest a nuanced answer: clarification helps, but only under the right conditions. Two structural factors shape this outcome. First, repairability defines a natural boundary: many parser errors cannot be resolved through local rewriting of the final utterance, even under a best-of-8 candidate pool, while a smaller subset remains consistently repairable. Second, effectiveness depends on selectivity. Within this subset, naive clarification often causes more harm than benefit, whereas a parser-aware RL policy can reduce regressions by learning when to abstain. Together, these findings suggest that the key challenge is not just generating better rewrites, but deciding when to intervene. This reframes clarification as a selective intervention problem. A key next step is \textbf{rewritability prediction}: estimating whether an instance is locally repairable before rewriting. More broadly, the prevalence of unrecoverable cases motivates moving beyond single-utterance rewriting toward context-level interventions, while the limits of black-box interaction suggest tighter integration between clarifier and parser. 

%% file: latex/8_limitations.tex
\section*{Limitations}
Our study examines last-utterance clarification within an incremental discourse parsing setting. While incremental discourse dependency parsing (DDP) is commonly adopted to model the streaming and real-time nature of dialogue~\citep{naim_towards_2025}, this setting represents only one possible modeling choice and may limit the generality of our findings. We note several limitations below.

First, our evaluation considers two representative parsers from the generative and discriminative families. Although these parsers reflect common design choices, other parsing paradigms—such as graph-based~\citep{afantenos-etal-2015-discourse,perret2016integer} and transition-based approaches~\citep{wang-etal-2017-two}—were not examined and may exhibit different sensitivities to clarification.

Second, our experiments are restricted to a single discourse formalism. It remains an open question whether similar patterns would be observed under alternative frameworks, such as RST~\citep{mann1988rhetorical} or PDTB~\citep{prasad-etal-2008-penn}.

Third, due to computational budget constraints and the scale of experimentation, we rely on a single type of open-weight LLM (e.g., Qwen3 in 8B and 14B). Results may differ with other model families or more recent frontier LLMs.


Finally, we investigate seven representative clarification strategies with manually selected conservative prompts optimized for meaning preservation, which capture common classes of linguistic edits but do not aim to exhaustively enumerate the space of possible clarification operations. We also do not directly compare against supervised clarification~\citep{fan-etal-2025-improving} or evaluate a learned rewrite-or-not classifier; both are natural baselines that our repair/regression framework enables as future work.

%% file: latex/9_ack.tex
\section*{Acknowledgments}
The authors wish to thank the anonymous reviewers and members of the OU NLP group for their
valuable feedback. This research was supported
by NSF ACCESS Allocation Request CIS250873, a compute grant from Modal.com, and a
GPU gift from NVIDIA Corporation.

%% file: latex/app-dataset.tex
\section{Parser-Agnostic Clarification and Dataset Statistics}
\label{sec:appendix-datasets}

\paragraph{Meaning Preservation Analysis.}
We report BERTScore between the original and rewritten utterances as a post-hoc diagnostic to verify that the clarifications are broadly meaning-preserving. These scores are computed on the test set, but are not used for prompt selection, model tuning, or evaluation, and do not affect the reported parsing results.  Their role is purely to confirm that the rewriting strategies do not \textbf{substantially} alter the underlying intent of the utterance. Since no model or prompt decisions are based on these scores, reporting them on the test set does not introduce evaluation leakage.


\begin{table}[H]
\centering
\scriptsize
\setlength{\tabcolsep}{3pt} 
\renewcommand{\arraystretch}{1.1}
\resizebox{\columnwidth}{!}{%
\begin{tabular}{llrr}
\toprule
Dataset & Method & Edit\% (n\_changed) & BERTScore-f1 \\
\midrule
\multirow{7}{*}{stac}
& \CLVZERO    & 9.76 (102)  & 0.9468  \\
& \CLVONE     & 20.67 (216) & 0.9369 \\
& \CLVTWO     & 21.63 (226) & 0.8907 \\
& \CLVTHREE   & 8.33 (87) & 0.9180 \\
& \CLVFOUR    & 46.22 (483) & 0.9227 \\
& \CLVMIX     & 28.23 (295) & 0.9367 \\
& \CLVFREE    & 49.86 (521) & 0.9285 \\
\midrule
\multirow{7}{*}{molweni}
& \CLVZERO    & 42.52 (1671) & 0.9721  \\
& \CLVONE     & 44.81 (1757) & 0.9653 \\
& \CLVTWO     & 22.43 (906) & 0.9511 \\
& \CLVTHREE   & 29.75 (1159) & 0.9485 \\
& \CLVFOUR    & 68.42 (2773) & 0.9620 \\
& \CLVMIX     & 69.06 (2711) & 0.9609 \\
& \CLVFREE    & 88.19 (3466) & 0.9451 \\
\midrule
\multirow{7}{*}{msdc}
& \CLVZERO    & 6.96 (342) & 0.9517 \\
& \CLVONE     & 12.31 (605) & 0.9445 \\
& \CLVTWO     & 14.63 (718) & 0.9379 \\
& \CLVTHREE   & 12.39 (607) & 0.9307 \\
& \CLVFOUR    & 44.83 (2231) & 0.9418 \\
& \CLVMIX     & 19.58 (963) & 0.9364 \\
& \CLVFREE    & 45.34 (2229) & 0.9331 \\
\bottomrule
\end{tabular}%
}
\caption{Average BERTScore over 3 clarification runs across three datasets via different clarification strategies.}
\label{tab:mb-bertscore}
\end{table}

%% file: latex/app-prompts.tex
\section{Selected Prompts for Parser-Agnostic Clarification}
\label{sec:appendix-prompts}
The following boxes show the selected prompts for each clarification strategies we studied in our paper.
\begin{tcblisting}{promptbox, title={\LVZERO Prompt}}
You are an expert dialogue rewriting assistant. Given the dialogue context and the last utterance, you will clarify the last utterance, but only when it is necessary and safe. Your task is to:

1. Identify if the last utterance has the following form-only issue:
- A high-confidence typo/misspelling with a single obvious correction (no guessing).

2. Then choose one of the following actions:
- If there is an above issue and the fix is safe, output the minimal allowed rewrite to fix only those issues.
- Otherwise, output the last utterance exactly unchanged.

Allowed edits:
- Fix high-confidence typos with one obvious correction.

Important:
- If you are not fully confident the edit is meaning-preserving, output the last utterance unchanged.
- Do NOT make any edits beyond the allowed edits above.
- Do NOT add new facts or information not explicitly in the context.
- Do NOT change meaning and intent.
- Do NOT change stance/tone/sentiment. Minimal casing/punctuation changes are allowed only as required by the allowed edits.

Show your reasoning only in <think> </think> tags. Final output of the last utterance must be inside <answer> </answer> as plain text (no speaker tag).

Perform clarification for the following dialogue context and the last utterance:

Context:
{context}

Last Utterance:
{utterance}
\end{tcblisting}

\begin{tcblisting}{promptbox, title={\LVONE Prompt}}
You are an expert dialogue rewriting assistant. Given the dialogue context and the last utterance, you will clarify the last utterance, but only when it is necessary and safe. Your task is to:

1. Identify if the last utterance has the following normalization issue: 
- It contains informal chat shorthand/abbreviation/slang that has ONE clear, widely accepted expansion in this context (no plausible alternative meanings).

2. Then choose one of the following actions:
- If there is an above issue and the fix is safe, output the minimal allowed rewrite to fix only those issues.
- Otherwise, output the last utterance exactly unchanged.

Allowed edits:
- Expand the shorthand token(s) to their single canonical expansion.
- Make the minimum number of expansions needed.

Important:
- If you are not fully confident the edit is meaning-preserving, output the last utterance unchanged.
- Do NOT make any edits beyond the allowed edits above.
- Do NOT add new facts or information not explicitly in the context.
- Do NOT change meaning and intent.
- Do NOT change stance/tone/sentiment. Minimal casing/punctuation changes are allowed only as required by the allowed edits.

Show your reasoning only in <think> </think> tags. Final output of the last utterance must be inside <answer> </answer> as plain text (no speaker tag).

Perform clarification for the following dialogue context and the last utterance:

Context:
{context}

Last Utterance:
{utterance}
\end{tcblisting}

\begin{tcblisting}{promptbox, title={\LVTWO Prompt}}
You are an expert dialogue rewriting assistant. Given the dialogue context and the last utterance, you will clarify the last utterance, but only when it is necessary and safe. Your task is to:

1. Identify if the last utterance has the following explicitness issue caused by ellipsis/fragmentation: 
- The last utterance is NOT a complete, stand-alone clause on its own (e.g., a fragment, short answer, or bare reply), AND
- The missing material (predicate / object / complement) is uniquely and explicitly recoverable from the dialogue context, AND
- Adding the missing words would NOT introduce new facts, new intentions, or new discourse relations beyond what is already entailed by the context.

2. Then choose one of the following actions:
- If there is an above issue and the fix is safe, output the minimal allowed rewrite to fix only those issues.
- Otherwise, output the last utterance exactly unchanged.

Allowed edits:
- Complete short answers/fragments by copying ONLY explicitly stated information from the context.
   - YES/NO answers: If the previous turn asks a yes/no question or proposes an action, you may expand "yes/no/okay/sure" into a full answer that repeats the proposition from context.
   - WH-answers (time/place/object): If the previous turn asks "when/where/which/what", you may expand a fragment (e.g., "tomorrow", "at 5", "the red one") into a complete clause that answers that question.
   - Justification-fragments: If the previous turn states the proposition being justified, you may attach it (e.g., "Because I was busy" -> "I can't make it because I was busy") ONLY if that proposition is explicitly stated in context.
- If the completion needs a subject, use only "I" (speaker of the last utterance) or "you" (addressing the other speaker) WHEN it is obvious from the previous utterances.

Important:
- If you are not fully confident the edit is meaning-preserving, output the last utterance unchanged.
- Do NOT make any edits beyond the allowed edits above.
- Do NOT add new facts or information not explicitly in the context.
- Do NOT change meaning and intent.
- Do NOT change stance/tone/sentiment. Minimal casing/punctuation changes are allowed only as required by the allowed edits.

Show your reasoning only in <think> </think> tags. Final output of the last utterance must be inside <answer> </answer> as plain text (no speaker tag).

Perform clarification for the following dialogue context and the last utterance:

Context:
{context}

Last Utterance:
{utterance}
\end{tcblisting}

\begin{tcblisting}{promptbox, title={\LVTHREE Prompt}}
You are an expert dialogue rewriting assistant. Given the dialogue context and the last utterance, you will clarify the last utterance, but only when it is necessary and safe. Your task is to:

1. Identify if the last utterance has the following coreference issue: 
- The last utterance contains a pronoun or deictic (e.g., it/this/that/these/those/he/she/they/him/her/them), AND
- The dialogue context contains exactly one unambiguous antecedent for it that is explicitly mentioned.

2. Then choose one of the following actions:
- If there is an above issue and the fix is safe, output the minimal allowed rewrite to fix only those issues.
- Otherwise, output the last utterance exactly unchanged.

Allowed edits:
- Replace the pronoun/deictic with its antecedent ONLY when the antecedent is explicit and unambiguous.
- Make the minimum number of replacements needed.

Important:
- If you are not fully confident the edit is meaning-preserving, output the last utterance unchanged.
- Do NOT make any edits beyond the allowed edits above.
- Do NOT add new facts or information not explicitly in the context.
- Do NOT change meaning and intent.
- Do NOT change stance/tone/sentiment. Minimal casing/punctuation changes are allowed only as required by the allowed edits.

Show your reasoning only in <think> </think> tags. Final output of the last utterance must be inside <answer> </answer> as plain text (no speaker tag).

Perform clarification for the following dialogue context and the last utterance:

Context:
{context}

Last Utterance:
{utterance}
\end{tcblisting}

\begin{tcblisting}{promptbox, title={\LVFOUR Prompt}}
You are an expert dialogue rewriting assistant. Given the dialogue context and the last utterance, you will clarify the last utterance, but only when it is necessary and safe. Your task is to:

1. Identify if the last utterance has the following discourse-marker issue:
- The last utterance's intended discourse move is clear from the context, but the relation to the immediately preceding utterance is left implicit (e.g., topic shift, continuation/addition, contrast, consequence/next-step), AND
- Adding a very short discourse marker would make this relation explicit WITHOUT changing the propositional content, intent, or tone, AND
- There is ONE single obvious discourse marker that fits (no plausible alternatives).

2. Then choose one of the following actions:
- If there is an above issue and the fix is safe, output the minimal allowed rewrite to fix only those issues.
- Otherwise, output the last utterance exactly unchanged.

Allowed edits:
- Add AT MOST ONE short discourse marker phrase, typically at the beginning of the last utterance, you may choose from the following set:
  Topic shift/return: "By the way,", "Anyway,", "Back to that,"
  Addition/continuation: "Also,", "And,", "In addition,"
  Contrast/correction: "But,", "However,", "Instead,"
  Result/next step: "So,", "Then,", "In that case,", "As a result,"
  Example/clarification: "For example,", "For instance,", "In other words,"
- Do not add any other words besides the marker (and required comma/spacing).
- The marker's casing must match the style of the last utterance.

Important:
- If you are not fully confident the edit is meaning-preserving, output the last utterance unchanged.
- Do NOT make any edits beyond the allowed edits above.
- Do NOT add new facts or information not explicitly in the context.
- Do NOT change meaning and intent.
- Do NOT change stance/tone/sentiment. Minimal casing/punctuation changes are allowed only as required by the allowed edits.

Show your reasoning only in <think> </think> tags. Final output of the last utterance must be inside <answer> </answer> as plain text (no speaker tag).

Perform clarification for the following dialogue context and the last utterance:

Context:
{context}

Last Utterance:
{utterance}
\end{tcblisting}

\begin{tcblisting}{promptbox, title={\LVMIX Prompt}}
You are an expert dialogue rewriting assistant. Given the dialogue context and the last utterance, you will clarify the last utterance, but only when it is necessary and safe. Your task is to:

1. Identify if the last utterance has ANY of the following issues:
- Form-only: a high-confidence typo/misspelling with one obvious correction.
- Normalization: informal shorthand/abbreviation/slang with ONE clear expansion in this context.
- Explicitness: ellipsis/fragment/underspecified answer where the missing words are uniquely recoverable from context and adding them is meaning-preserving.
- Coreference: a pronoun/deictic with exactly one explicit, unambiguous antecedent in context.
- Discourse marker: the connection to the immediately preceding utterance is clear but implicit, and exactly one obvious short discourse marker would make it explicit without changing meaning.

2. Then choose one of the following actions:
- If there is an above issue and the fix is safe, output the minimal allowed rewrite to fix only those issues.
- Otherwise, output the last utterance exactly unchanged.

Allowed edits:
- Fix high-confidence typos (one obvious correction).
- Expand shorthand tokens to their single canonical expansion.
- Complete ellipsis/fragments ONLY when missing material is explicitly recoverable from context (no guessing).
- Replace pronouns/deictics with their explicit unambiguous antecedent (minimum replacements).
- Add AT MOST ONE discourse marker phrase at the beginning, you may choose from:
  Topic shift/return: "By the way,", "Anyway,", "Back to that,"
  Addition/continuation: "Also,", "And,", "In addition,"
  Contrast/correction: "But,", "However,", "Instead,"
  Result/next step: "So,", "Then,", "In that case,", "As a result,"
  Example/clarification: "For example,", "For instance,", "In other words,"
  The marker's casing must match the style of the last utterance.

Important:
- If you are not fully confident the edit is meaning-preserving, output the last utterance unchanged.
- Do NOT make any edits beyond the allowed edits above.
- Do NOT add new facts or information not explicitly in the context.
- Do NOT change meaning and intent.
- Do NOT change stance/tone/sentiment. Minimal casing/punctuation changes are allowed only as required by the allowed edits.

Show your reasoning only in <think> </think> tags. Final output of the last utterance must be inside <answer> </answer> as plain text (no speaker tag).

Perform clarification for the following dialogue context and the last utterance:

Context:
{context}

Last Utterance:
{utterance}
\end{tcblisting}

\begin{tcblisting}{promptbox, title={\LVFREE Prompt}}
You are an expert dialogue rewriting assistant. Given the dialogue context and the last utterance, you will clarify ONLY the last utterance to be clearer and easier to understand, while preserving the original meaning and intent. Your task is to:

1. Identify if the last utterance has ambiguity or implicitness that could cause misunderstanding (e.g., typo, abbreviations, slang, vague references, incomplete phrasing, etc).
2. Then choose one of the following actions:
- If yes and you can safely preserve meaning, rewrite the last utterance to be clearer.
- Otherwise, output the last utterance exactly unchanged.

Important:
- Do NOT add new facts or information not explicitly in the context.
- Do NOT change meaning and intent.
- Do NOT change stance/tone/sentiment.
- The rewrite must sound like a natural next turn after the given context and not disrupting the dialogue flow. 
- Make only the minimal changes needed for clarity. 

Show your reasoning only in <think></think> tags. Final output must be inside <answer></answer> as plain text (no speaker tag).

Perform clarification for the following dialogue context and the last utterance:

Context:
{context}

Last Utterance:
{utterance}
\end{tcblisting}

%% file: latex/app-rl.tex
\section{RL-based Parser-Aware Clarification}
\label{sec:appendix-rl}

We now describe how we train the rewriting policy $\pi_\phi$ with reinforcement learning to improve the frozen parser's output. At a high level, each training example is treated as a one-step episode: the policy rewrites the current utterance, the parser returns a structured prediction, and a scalar reward is computed from the change in discourse quality.

\paragraph{RL training with GRPO. } For each dialogue $x = (u_1,\dots,u_T)$ and position $t$, we define a state $s_t$ as the dialogue prefix and current utterance, $s_t = (u_1,\dots,u_t)$. The policy $\pi_{\phi}$ takes $s_t$ as input and generates a textual output
that contains both reasoning and the clarified utterance. We form a clarified prefix by replacing $u_t$ with $\tilde{u}_t$ and run the frozen parser on both the original and clarified prefixes. The environment then returns a scalar reward $r$ computed from these two outputs and the gold discourse annotation.

We optimize $\pi_\phi$ with GRPO: for each state we sample a group of candidate clarifications, query the frozen parser on each, and compute each candidate's advantage by normalizing its reward within the sampled group rather than learning a value network. To keep the policy close to the base LLM and preserve fluent, semantically plausible clarifications, we include a KL penalty that discourages large deviations from the initial policy distribution. Throughout training, the parser remains frozen and is only used to produce discrete discourse graphs for reward computation.

\paragraph{Reward Design.} To facilitate effective RL training, we design verifiable reward signals tailored to the discourse parsing task. We decompose the reward into two parts: \emph{format} rewards and \emph{parsing} rewards.

We encourage models to follow a strict response format of the form \texttt{<think>...</think> <answer>...</answer>}. The output must contain exactly one \texttt{<think>} block and one \texttt{<answer>} block in this order, with no stray text outside the tags. If the format is correct, we assign a format reward $r_{\text{fmt}} = +1$; otherwise we assign a penalty $r_{\text{fmt}} = -2$ and \emph{do not} compute any parsing reward for this episode (i.e., we set $r_{\text{parse}} = 0$). The total reward is
\begin{equation}
    r = r_{\text{fmt}} + r_{\text{parse}}.
\end{equation}

When the format is correct, we compute a parsing reward based on how the clarification changes the frozen parser's prediction for the current step. Let $c_{\text{orig}}$ indicate whether the parser's prediction on the original utterance is fully correct (both link and relation match the gold), and $c_{\text{rew}}$ indicate whether the prediction on the clarified utterance is fully correct. We also define $p_{\text{rew}}$ to indicate a \emph{partial match} on the clarified utterance: the parser attaches the link to the correct target but predicts an incorrect relation label. The parsing reward is then

\begin{equation}
r_{\text{parse}} =
    \begin{cases}
    +2 & \text{if } \neg c_{\text{orig}} \land c_{\text{rew}} \\
    -2 & \text{if } c_{\text{orig}} \land \neg c_{\text{rew}} \\
    +1 & \text{if } \neg c_{\text{orig}} \land \neg c_{\text{rew}} \land p_{\text{rew}} \\
    0  & \text{otherwise.}
    \end{cases}
\end{equation}

Thus, the clarifier receives a large positive reward when it fixes an originally incorrect prediction, a large negative reward when it breaks an originally correct one, and a smaller positive reward when it at least corrects the attachment while still predicting the wrong relation. Neutral or ambiguous changes receive zero parsing reward. Together with the format term, this encourages the policy to produce well-structured outputs and clarifications that genuinely help the frozen parser on its structured discourse objective.

%% file: latex/app-parsers.tex
\section{Experiment Setup}
\label{sec:appendix-parsers}

\subsection{LLM-based SFT Parser}
\label{ssec:sft-parser}
\autoref{tab:finetuning} describes hyperparameters we used when finetuning Qwen3-8B (\url{https://huggingface.co/Qwen/Qwen3-8B}).

\begin{table}[h]
    \centering
    \begin{tabular}{lc}
        \hline
         Hyperparameter & Value \\
         \hline
         batch size & 16 \\
         optimizer & AdamW \\
         learning rate & 5e-5 \\
         weight decay & 0.01 \\
         lora dropout & 0.05 \\
         lora rank & 64 \\
         lora alpha & 128 \\
         \hline
    \end{tabular}
    \caption{Hyperparameters of finetuning Qwen3-8B on Stac, Molweni, MSDC}
    \label{tab:finetuning}
\end{table}

\subsection{SDDP Parser}
\label{ssec:sddp-parser}
We use the official code (\url{https://github.com/chijames/structured_dialogue_discourse_parsing}) released by the SDDP authors and train the model following their default hyperparameter settings.

\subsection{GRPO-based Parser-aware Clarification}
\label{ssec:grpo-parser}
The follow table describes hyperparameters we used for RL training. We use the Instruct-tuned version of Qwen2.5-7B for all experiments.

\begin{table}[h]
    \centering
    \begin{tabular}{lc}
        \hline
         Hyperparameter & Value \\
         \hline
         epoch & 8 \\
         max prompt length & 1024 \\
         max response length & 768 \\
         batch size & 96 \\
         rollout & 8 \\
         learning rate & 1e-6 \\
         sampling temperature & 0.7 \\
         lora rank & 32 \\
         lora alpha & 32 \\
         \hline
    \end{tabular}
    \caption{Hyperparameters of GRPO training.}
    \label{table4}
\end{table}

\paragraph{Training Details. } All RL models are trained using the verl (\url{https://github.com/volcengine/verl}) framework and are conducted on 4 NVIDIA RTX 6000 96GB GPUs. 

%% file: latex/app-analysis.tex
\section{More Analysis on Results}
\label{sec:appendix-cases}

\begin{table*}[t]
\centering\small\setlength{\tabcolsep}{4pt}
\begin{tabular}{llp{2.4cm}p{2.8cm}lll}
\toprule
\textbf{Eff.} & \textbf{Strat.}
  & \textbf{Original} & \textbf{Rewrite}
  & \textbf{Gold} & \textbf{Before} & \textbf{After} \\
\midrule
fix   & L1 & \textit{yeah}
            & \textit{yes}                  & QAP   & ACK   & QAP  \\[2pt]
fix   & L1 & \textit{go nz}
            & \textit{go New Zealand}       & RES   & CONT  & RES  \\
\midrule
regr. & L1 & \textit{lol}
            & \textit{laughing out loud}    & CMT   & CMT   & ELAB \\[2pt]
regr. & L0 & \textit{not like peurto rico}
            & \textit{not like Puerto Rico} & CONTR & CONTR & CMT  \\
\bottomrule
\end{tabular}
\caption{Surface-form changes that flip the Qwen3-8B parser output despite
  carrying no meaning difference for a human reader (STAC test).
  Before/After~= parser prediction on original/rewritten utterance.
  QAP~= Question-answer\_pair; ACK~= Acknowledgement;
  CONT~= Continuation; RES~= Result;
  CMT~= Comment; ELAB~= Elaboration; CONTR~= Contrast.}
\label{tab:l0l1-surface}
\end{table*}

Our results point to a central limitation of input-side intervention for frozen dialogue discourse parsers. The main bottleneck is not simply generating better rewrites, but aligning the scope of intervention with the source of parser error and identifying when intervention is warranted. Across parser-agnostic strategies, clarification rarely yields net gains because the same edits that create headroom for repairs also introduce regressions, often overwhelming the benefit of the repaired cases. This pattern holds across datasets and across both generative and discriminative parsers, suggesting that the instability is not tied to one particular model family, but to the mismatch between local surface editing and discourse decisions that depend on broader structured context.

\paragraph{Why is the selectivity problem so hard?} The regression patterns in Section~\ref{ssec:parser-agnostic-strateiges} reveal that parser behavior is not a smooth function of surface form. Even semantically similar rewrites of the same utterance can produce opposite parser outcomes: one may correct an attachment error, while another creates a new one. These outcomes are not reliably predictable from surface form alone. Such unpredictability distinguishes discourse parsing from tasks like retrieval or translation where improvements in input quality often yield more gradual gains in output quality. The RL clarifier's behavior in Section~\ref{ssec:rq3-rl-parser-aware} reflects this directly: the policy learns \textit{when not to act} but struggles to learn \textit{how to act well}. 

Negative signal from regressions is frequent and consistent, teaching the policy to be conservative. But positive signal from repairs is sparse and noisy — not because the policy generates poor rewrites, but because most errors fall in the unrepairable 80\%, meaning that even high-quality rewrites receive zero reward. The policy cannot distinguish a good rewrite that fails due to structural unrepairability from a bad rewrite that fails on its own terms. This asymmetry explains why GRPO training converges toward selective abstention rather than toward better clarification. Recent work has increasingly sophisticated mechanisms for densifying rewards and improving credit assignment in exactly this kind of outcome-reward-through-a-frozen-model setup, including gradient attribution from the frozen judge, implicit process rewards derived from outcome labels, and entropy-weighted per-token credit assignment within GRPO. Our analysis suggests none of these would help in our setting because they do not address the scope mismatch described in Section~\ref{ssec:rq3-rl-parser-aware}, where 80\% of errors originate outside the intervention window and are therefore unreachable regardless of how finely credit is attributed. The ceiling is structural rather than a consequence of coarse training signal.

\paragraph{Why does surface clarification fail for discourse parsing?} Our results provide empirical support for a theoretically grounded explanation. In SDRT, discourse relations are determined through semantic inference over structured context, not through surface pattern matching. A parser trained on naturalistic dialogue learns to exploit subtle interactions between context and the current utterance, including implicit signals such as the absence of a connective, a pronoun left unresolved, or a fragmentary form that signals continuation. When a clarifier makes these signals explicit, it does not simply add information; it replaces one surface realization with another that falls outside the parser's learned distribution. This explains why connective insertion (L4-CONN) is consistently the most harmful strategy: it directly targets the discourse-level signal, but the parser was never trained on utterances with explicit connectives in positions where the original data left them implicit.

\paragraph{What should the community pursue instead?} Our findings point to three directions. First, the 80\% unreachable ceiling suggests that multi-utterance or context-level rewriting may be necessary to address errors that originate outside the last utterance, such as long-range dependencies and context-side ambiguity. Second, the selectivity bottleneck motivates rewritability prediction as a standalone task: a lightweight classifier that estimates whether a given utterance is likely to benefit from rewriting before any rewrite is attempted. Such a classifier could be trained on the repair/regression labels generated by our experimental framework, and it could condition on parser uncertainty signals (e.g., entropy over candidate relations) to improve triggering precision. Third, our results suggest that tighter integration between clarifier and parser may be needed for consistent gains. Rather than treating the parser as a fully black-box system, approaches that expose partial parser state (attention distributions, candidate rankings) to the clarifier could enable more targeted interventions. This would move beyond the strict frozen-parser constraint we study here, but our findings suggest that this constraint is precisely what limits the approach.

More broadly, our results carry a cautionary message for agentic AI pipelines that compose upstream rewriters with frozen downstream tools. While this modular pattern works well for tasks where surface form directly determines output quality (retrieval, translation), it is less effective for tasks governed by semantic inference, where the downstream tool's decisions depend on implicit distributional signals that rewriting may disrupt. Evaluating such pipelines requires the kind of repair-vs-regression accounting we employ here, as aggregate metrics can mask harmful failure modes.